\pdfoutput=1
\documentclass[runningheads]{llncs}
\usepackage{graphicx}
\usepackage{tikz}
\usepackage{comment}
\usepackage{amsmath,amssymb} 
\usepackage{color}
\usepackage{algorithm}
\usepackage{algorithmic}

\usepackage[accsupp]{axessibility}  

\usepackage[width=122mm,left=12mm,paperwidth=146mm,height=193mm,top=12mm,paperheight=217mm]{geometry}

\begin{document}
\pagestyle{headings}
\mainmatter

\title{Hand-Assisted Expression Recognition Method from Synthetic Images at the Fourth ABAW Challenge} 

\author{Xiangyu Miao\and Jiahe Wang\and Yanan Chang\and Yi Wu \and Shangfei Wang}
\institute{University of Science and Technology of China, Hefei, China\\
\email{\{ mxy3369, pia317, cyn123, wy221711\}@mail.ustc.edu.cn}\\
\email{sfwang@ustc.edu.cn}}
\titlerunning{Hand-Assisted Expression Recognition Method}

\maketitle

\begin{abstract}
Learning from synthetic images plays an important role in facial expression recognition task due to the difficulties of labeling the real images, and it is challenging because of the gap between the synthetic images and real images. The fourth Affective Behavior Analysis in-the-wild Competition raises the challenge and provides the synthetic images generated from Aff-Wild2 dataset. In this paper, we propose a hand-assisted expression recognition method to reduce the gap between the synthetic data and real data. Our method consists of two parts: expression recognition module and hand prediction module. Expression recognition module extracts expression information and hand prediction module predicts whether the image contains hands. Decision mode is used to combine the results of two modules, and post-pruning is used to improve the result. F1 score is used to verify the effectiveness of our method.
\end{abstract}

\section{Introduction}
Facial expression recognition has received widespread attention in human computer interaction. real images are used for facial expression recognition in most cases, and sometimes synthetic images are used as auxiliary images. However, few methods try to train expression recognition only on synthetic images. Labeling facial expression images is time-consuming and labor-intensive, and even for experts, it is difficult to label all images correctly. Learning from synthetic images plays an important role in facial expression recognition task due to the difficulties of labeling the real images. In the fourth Affective Behavior Analysis in-the-wild Competition\cite{kollias2022abaw}, learning from synthetic data challenge is raised.

For this challenge, around 300K synthetic images have been generated that contain annotations in terms of the 6 basic facial expressions (anger, disgust, fear, happiness, sadness, surprise). Using synthetic images makes it easier to obtain data, but it also increases the difficulty of facial expression recognition task. Most images are generated by zooming in on certain expression characteristics such as opening mouth, frowning and so on. These characteristics satisfy some micro-expression prior knowledge, but there is still a certain gap with the real expression. In real images, each expression is various, but the characteristics of each expression in the synthetic image are simple. This leads to increased difficulty in learning expression information from synthetic images. 

In this paper, we propose a hand-assisted expression recognition method for expression recognition from synthetic images. Our method consists of two parts: expression recognition module and hand prediction module. Specifically, expression recognition module is pretrained on the Multi-PIE dataset \cite{gross2010multipie} to keep the prior knowledge of expressions in real images. This module is fine-tuned on the synthetic dataset to realize domain adaption. What's more, from Emotion \cite{emotion} there is a certain relationship between hand movements and expressions, so hand prediction module is used to learn the presentations of fear. Decision mode is used to combine the results of two modules, and post-pruning is used to improve the result.

\section{Proposed Method}

\subsection{Expression Recognition Module}
The authors \cite{he2016renet} propose the residual nets (ResNet) as an effective solution in a deep learning network. We use the ResNet50 as the backbone of the expression recognition module. Firstly, this module is pretrianed on the Multi-PIE dataset to learn real expression features. Cross entropy loss is used for pretraining. The first 80\% of the residual blocks are fixed for keeping the real expression feature. The rest are fine-tuned on the synthetic images. Cross entropy loss $L_c$ is used for training, which is formulated as:
\begin{equation}
    L_c = -\sum ylog(\hat{y}).
\end{equation}

\subsection{Hand Prediction Module}
In the book Emotion \cite{emotion}, fear is accompanied by a defense mechanism for the startle reflex, and this is also widespread in the validation set. These images are quite different from the generated fear images. All of the fear images are generated by enlarging the eyes and opening the mouth of normal fear images. This method amplifies the expression of fear on expressions, but ignores the relationship between action and psychology in behavioral psychology. This lead to a huge gap between real images and synthetic images.

Here hand prediction module is introduced to reconstruct this relationship. Covering faces with hands is seen as a sign of fear, and hand prediction module will check whether the images contain hands. To better detect the hands, Sobel operator is used to extract the edge information of the images. Then the extracted edge information is input into the Resnet50 for binary classification. Cross entropy loss is also used for training.

\subsection{Decision Mode}

It's not just fear that causes the hands to appear in the image. In some cases, other emotions will also lead to this situation, like happiness. By analyzing the predicted results, due to the small eyes or the opened mouth, fear images with hands are mainly classified into sadness or surprise without the hand prediction module. Therefore, if the result of hand prediction module is true, the image will be classified into fear. However, the happiness images will be classified incorrectly because in happiness images, hands also appears. Happiness itself is an easily recognizable expression, and its similarity with fear is only the hand, so we assume that predicted result of happiness is always correct. Thus, the algorithm can be written as Algorithm \ref{alg:1}:
\begin{algorithm}
	\renewcommand{\algorithmicrequire}{\textbf{Input:}}
	\renewcommand{\algorithmicensure}{\textbf{Output:}}
	\caption{The recognition process of our method}
	\label{alg:1}
	\begin{algorithmic}[1]
		\REQUIRE the image $x$ from the testing set $X$, trained expression recognition module $E$, trained hand prediction module $H$ and the Sobel operator $S$
		\ENSURE the predicted label $y$ of the input image
		\STATE Given the input image $x$, calculate the edge map $x'=S(x)$ 
		\STATE Input the image $x$ into $E$ to get the expression label $y_e = E(x)$
		\STATE Input the edge map $x'$ into $H$ to get the Hand label $y_h = H(x')$
		\IF{$y_h$ and $y_e \neq$ Happiness}
		\STATE \textbf{return} Fear
		\ELSE
		\STATE \textbf{return} $y_e$
		\ENDIF
	\end{algorithmic}  
\end{algorithm}

\section{Experiments}

\subsection{Dataset}
The synthetic Aff-Wild2 dataset \cite{kollias2022abaw,kollias2021distribution,kollias2021affect,kollias2020deep,kollias2020va,kollias2019expression,kollias2019deep,kollias2018photorealistic,kollias2017recognition,zafeiriou2017aff} contains about 300k images annotated on 6 basic expression. There are 18286 anger images, 15150 disgust images, 10923 fear images, 73285 happiness images, 144631 sadness images and 14976 surprise images. The image size is 128 $\times$ 128. For hand prediction module training, We manually label the presence or absence of hand labels on the synthetic dataset.

\subsection{Training Details}
Both expression recognition module and hand prediction module are trained on the Pytorch framework. Adam optimization \cite{kingma2014adam} is used to update the weights. The learning rate is $5e^{-5}$ and weight decay is $e^{-4}$. Our models are trained with the epoch of 20 and saves the best performance on the validation set. Our expression recognition module is firstly trained on the Multi-PIE dataset and then finetuned on the synthetic Aff-Wild2 dataset. Weighted sampling is used to mitigate the effects of class imbalance. In this challenge, the final result is evaluated across the average F1 score of 6 emotion categories:
\begin{equation}
    F_1^{final} = \frac{\sum F_1^{expression}}{6},
\end{equation}
where $F_1^{expression}$ is $F_1$ score of each expression.

\subsection{Results}
The results on the validation by $F_1$ score are shown in Table \ref{tab:r1}. Even with the introduction of a pretrained model with real data, the F1 score is only improved by 0.7\%. This represents a bias in expression that exists across real pictures between the datasets. Pretraining process improves the module, but due to the bias, it is not effective. After adding the hand prediction module and setting the results of this module as fear, the results improves 4.4\%. This is because hand prediction reduces the gap between the validation set and the training set. In training set, the character of fear is that eyes are opened while in the validation set, the eyes are closed and hands cover the face. By using the hand prediction module, the gap reduces. What's more, using post-pruning, the $F_1$ score increases 2.6\%. If simply thinking that an image with hands is fear, there will be lots of mistakes in the happiness expression. Through post-pruning, the influence of hand prediction module on predicting happiness is eliminated, so the result is further improved. 
\begin{table}
    \centering
    \begin{tabular}{|c|c|}
        \hline
        Model & Validation(\%) \\
        \hline
        baseline(Resnet50) \cite{he2016renet} & 62.3 \\
        \hline
        baseline + pretraining  & 63.0 \\
        \hline
        baseline + pretraining + Hand  &   67.4 \\
        \hline
        baseline + pretraining + Hand + post-pruning &   70.0 \\
        \hline
    \end{tabular}
    \caption{Expression recognition results on the validation dataset}
    \label{tab:r1}
\end{table}

\section{Conclusions}

In this paper, we propose a hand-assisted expression recognition method for expression recognition from synthetic images. Our method consists of two parts: expression recognition module and hand prediction module. expression recognition module extracts expression information and hand prediction module predicts whether the image contains hands. Decision mode is used to combine the results of two modules, and post-pruning is used to improve the result.

%
%
\bibliographystyle{splncs04}
\bibliography{egbib}
\end{document}